\documentclass[sigconf]{acmart}

\usepackage{amsmath}
\usepackage{graphicx}
\usepackage{booktabs}
\usepackage{multirow}
\usepackage[table]{xcolor}
\definecolor{oursgray}{gray}{0.95}
\definecolor{oursgreen}{RGB}{232,245,233}
\definecolor{oursblue}{RGB}{242,246,252}
\newcommand{\dec}[1]{\textcolor{green!60!black}{($\downarrow$#1\%)}}
\newcommand{\inc}[1]{\textcolor{red!75!black}{($\uparrow$#1\%)}}
\newcommand{\abhead}[2]{\parbox[c][2.4em][c]{#1}{\centering #2}}

\setcopyright{cc}
\setcctype{by}

\copyrightyear{2026}
\acmYear{2026}
\acmDOI{10.1145/3767308.3835103}
\acmConference[MM '26]
  {Proceedings of the 34th ACM International Conference on Multimedia}
  {November 10--14, 2026}
  {Rio de Janeiro, Brazil}
\acmBooktitle{Proceedings of the 34th ACM International Conference on
  Multimedia (MM '26), November 10--14, 2026, Rio de Janeiro, Brazil}
\acmISBN{979-8-4007-2213-4/2026/11}
\begin{document}

\title[GSTEP: Global Spatio-Temporal Density Pruning]{GSTEP: Global Spatio-Temporal Density-Driven Visual Token Pruning for Efficient Video Large Language Models}

\author{Mengjie Zhang}
\authornote{These authors contributed equally to this work.}
\orcid{0009-0001-9648-6855}
\email{zhangmengjie@mail.ustc.edu.cn}

\author{Qihui Zhu}
\authornotemark[1]
\orcid{0009-0000-8765-6982}
\email{qh.zhu@mail.ustc.edu.cn}

\author{Tao Zhang}
\authornotemark[1]
\orcid{0009-0009-5391-5048}
\email{zhangtaolqy@mail.ustc.edu.cn}
\affiliation{%
  \institution{University of Science and Technology of China}
  \city{He Fei}
  \country{China}}

\author{Shuangwu Chen}
\orcid{0000-0003-2817-9738}
\email{chensw@ustc.edu.cn}
\author{Huihuang Qin}
\orcid{0009-0005-7860-4896}
\email{huihqin@mail.ustc.edu.cn}
\author{Yu Guo}
\orcid{0000-0003-2571-0247}
\email{yukariguo@mail.ustc.edu.cn}
\affiliation{%
  \institution{University of Science and Technology of China}
  \city{He Fei}
  \country{China}}

\author{Shenghao Ye}
\orcid{0009-0001-8173-8132}
\email{ssh0321y@mail.ustc.edu.cn}
\author{Zijian Wen}
\orcid{0009-0001-9015-6732}
\email{wzj20020304@mail.ustc.edu.cn}
\affiliation{%
  \institution{University of Science and Technology of China}
  \city{He Fei}
  \country{China}}

\author{Yunpeng Hou}
\authornote{Corresponding author.}
\orcid{0000-0001-7216-9022}
\email{hyp314@mail.ustc.edu.cn}
\author{Dong Jin}
\orcid{0000-0003-4026-8338}
\email{kingdon@mail.ustc.edu.cn}
\affiliation{%
  \institution{Institute of Artificial Intelligence, Hefei Comprehensive National Science Center}
  \city{He Fei}
  \country{China}}

\author{Xiaobin Tan}
\orcid{0000-0001-7489-2839}
\email{xbtan@ustc.edu.cn}
\author{Huasen He}
\orcid{0000-0001-9963-019X}
\email{hehuasen@ustc.edu.cn}
\author{Jian Yang}
\orcid{0000-0002-7329-4738}
\email{jianyang@ustc.edu.cn}
\affiliation{%
  \institution{University of Science and Technology of China}
  \city{He Fei}
  \country{China}}

\renewcommand{\shortauthors}{Mengjie Zhang et al.}

\begin{abstract}
Video large language models (VideoLLMs) exhibit exceptional video understanding performance, yet their inference costs remain prohibitively high due to the massive volume of visual tokens required, especially for long video understanding. Visual token pruning is a promising paradigm for mitigating such inference costs by eliminating redundant tokens across video frames. However, most existing token pruning methods adopt a segment-level pruning strategy, where videos are partitioned into isolated segments and tokens are selected independently within each segment. Such designs may cause information-dense segments to discard tokens that are seemingly non-salient from a local perspective but remain critical from a global one. To address this issue, we propose GSTEP (Global Spatio-Temporal Density Pruning), a plug-and-play pruning framework that models video as a continuous spatio-temporal information flow. GSTEP constructs a token-level spatio-temporal density by combining a continuous temporal density, obtained from a smoothed centered frame-level change signal, with intra-frame spatial density, and then performs global token sampling by jointly balancing information density and coverage. Extensive experiments on multiple VideoLLMs and public benchmarks demonstrate that GSTEP consistently achieves strong accuracy-efficiency trade-offs and generalizes well across model architectures and evaluation settings. On LLaVA-OneVision-7B, GSTEP prunes 75\% of visual tokens, preserves up to 100.2\% of the original average performance across benchmarks, and achieves a 1.17$\times$ end-to-end speedup. Code is available at \url{https://github.com/yeluoy/GSTEP}.
\end{abstract}

\ccsdesc[500]{Computing methodologies~Artificial intelligence}
\keywords{Video Large Language Models, Visual Token Pruning, Spatio Temporal Density}

\maketitle

\begin{figure}[t]
    \centering
    \includegraphics[width=\linewidth]{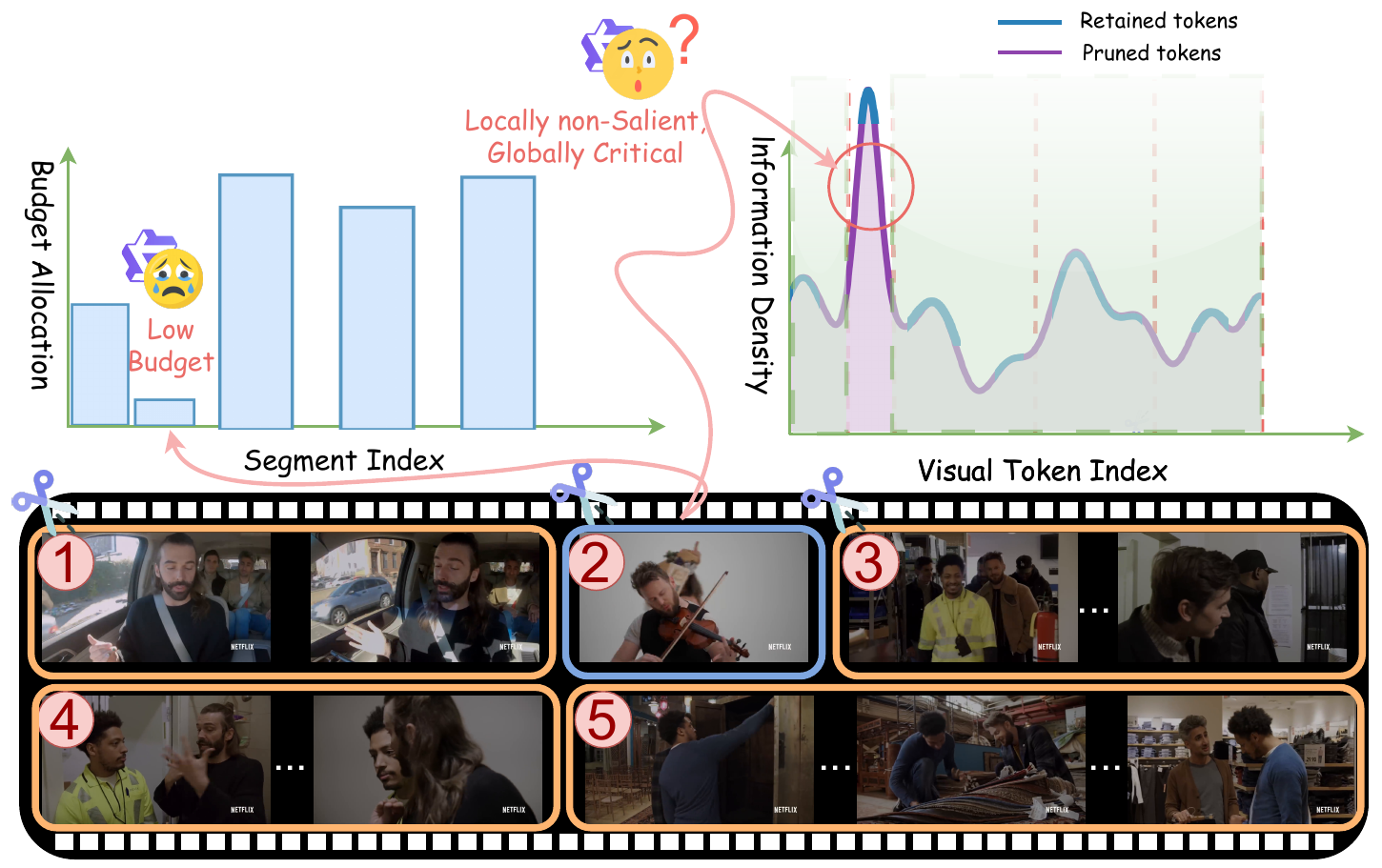}
    \caption{Limitations of existing segment-level pruning methods. Segment 2 is allocated a low budget due to its short duration. Tokens that are non-salient locally but globally critical are discarded.}
    \label{fig:global_pruning_advantage}
\end{figure}
\section{Introduction}
Video large language models (VideoLLMs) have exhibited strong video understanding capabilities by combining visual perception with language reasoning~\cite{video_llava,llava_ov,videochat,pllava,flamingo,blip2}. 
However, videos contain far more visual tokens than static images, driving up inference expenses remarkably~\cite{vit,vivit,moviechat,longvila}.
To tackle this challenge, recent studies have explored visual token pruning for VideoLLMs to mitigate the inherent redundancy in consecutive video frames~\cite{fastvid,vidcom2,prunevid,dycoke}.
Existing methods~\cite{prunevid,fastvid,dycoke,framefusion} generally adopt a budget allocation paradigm: a video is first partitioned into isolated segments of varying durations, after which an unequal token budget is assigned to each segment, and visual token pruning is accordingly conducted on each segment based on its local information density distribution. 
Such segment-level pruning methods impose hard boundaries that can limit pruning quality in two ways. 
(1) Rigid budget allocation starves short but crucial segments. As shown in Figure \ref{fig:global_pruning_advantage}, although Segment 2 has a high information density, it is assigned a low token budget due to its short duration, which inevitably leads to the loss of critical tokens. 
(2) A bounded local budget forces information-dense segments to discard tokens that are seemingly non-salient from a local perspective but remain critical from a global one. 

The above limitations reveal a pivotal insight that a video should be treated as a continuous spatio-temporal information flow rather than a set of isolated segments, enabling visual token pruning from a global perspective.
In light of this insight, we propose \textbf{GSTEP}, a \textbf{G}lobal \textbf{S}patio-\textbf{T}emporal d\textbf{E}nsity-driven visual token \textbf{P}runing method for VideoLLMs. 
Specifically, GSTEP first constructs a continuous temporal density by smoothing a centered frame-level change and then combines this temporal signal with intra-frame spatial density to form a token-level spatio-temporal density.
Based on this unified density representation, GSTEP performs global token sampling by jointly balancing information density and coverage, thereby preserving semantically critical transitions and continuous semantic flow under limited token budgets.
To evaluate the generalization capability of \textsc{GSTEP}, we conduct experiments on multiple representative VideoLLMs, including LLaVA-OneVision~\cite{llava_ov}, Qwen2.5-VL~\cite{qwen2_5_vl}, and Qwen3-VL~\cite{qwen3_vl}. We further validate its effectiveness on diverse public benchmarks—MVBench~\cite{mvbench}, LongVideoBench~\cite{longvideobench}, MLVU~\cite{mlvu}, VideoMME~\cite{videomme}, and EgoSchema~\cite{egoschema}—which span a wide range of video lengths, content complexity, and reasoning skills, enabling a comprehensive and challenging evaluation. Across these settings, \textsc{GSTEP} consistently achieves favorable accuracy-efficiency trade-offs. For example, on LLaVA-OneVision-7B, \textsc{GSTEP} prunes \textbf{75\%} of the visual tokens while maintaining up to \textbf{100.2\%} of the original average performance across benchmarks, achieving a \textbf{1.17$\times$} end-to-end speedup and a \textbf{1.61$\times$} speedup in the LLM stage. These results demonstrate that \textsc{GSTEP} generalizes well across different model backbones and benchmark distributions, providing robust token compression without requiring model-specific tuning.

The main contributions of this work are three-fold.
\begin{itemize}
    \item  
    By revealing that segment-level pruning induces starvation of short segments and discard of critical tokens, we present a key insight that video token pruning ought to be performed from a global, video-wide perspective.
    \item We propose GSTEP, a global spatio-temporal density-driven pruning framework that treats videos as a continuous spatio-temporal information stream instead of isolated segments, and  prunes redundant tokens according to the global spatio-temporal density.
    \item Extensive experiments on multiple VideoLLMs and benchmarks demonstrate that GSTEP consistently achieves strong performance retention and efficiency gains, while generalizing well across architectures and evaluation settings.

\end{itemize}

\section{Related Work}
\subsection{Video Large Language Models}
Recent video large language models~\cite{video_llava,llava_ov,qwen2_5_vl,qwen3_vl,internvl,videollama,minigpt4v,video-xl,videochat,pllava} extend multimodal large language models~\cite{flamingo,blip2,instructblip,visual} from static images to temporally rich video inputs, enabling open-ended video understanding, reasoning, and question answering~\cite{videosurvy,videosurvy2}. Representative models, such as Video-LLaVA~\cite{video_llava}, LLaVA-OneVision~\cite{llava_ov}, Qwen2.5-VL~\cite{qwen2_5_vl}, and Qwen3-VL~\cite{qwen3_vl}, typically follow a unified pipeline that combines a visual encoder, a projector or merger for modality alignment, and a large language model for cross-modal reasoning and autoregressive generation~\cite{visual,minigpt4,videollm,videosurvy2}. This design has led to strong performance on a wide range of video understanding benchmarks~\cite{mvbench,videomme,video_chatgpt,egoschema,mlvu,longvideobench}.

Despite their impressive capabilities, VideoLLMs also introduce a substantial efficiency challenge. Compared with image inputs, videos produce significantly more visual tokens due to the additional temporal dimension, which greatly increases memory consumption and computational cost in the language model, especially during the prefill stage~\cite{vit,vivit,moviechat,longvila}. This issue becomes even more severe for long-video scenarios, where large amounts of temporally redundant visual tokens are often fed into the LLM. Consequently, reducing redundant visual tokens while preserving reasoning performance has become an important direction for improving the efficiency of VideoLLMs, motivating a growing body of research on visual token pruning and compression~\cite{efficient,dynamicvit,tome,fastv,pdrop,framefusion,sparsevlm,multi,fitprune,pact,topv}.

\subsection{Visual Token Pruning and Compression}

A large body of work has explored visual token pruning and compression to improve the efficiency of vision and multimodal transformers~\cite{efficient,efficient2}. Early studies mainly focused on image-based or generic MLLM settings, where the primary goal is to reduce spatial redundancy or remove visually unimportant tokens~\cite{dynamicvit,evit,fastv,adaptive,tome}. Representative examples include token merging methods such as ToMe~\cite{tome}, attention-based pruning methods such as FastV~\cite{fastv} and VisionZip~\cite{visionzip}, and progressive pruning approaches such as PDrop~\cite{pdrop} and CDPruner~\cite{cdpruner}. These methods typically rely on token similarity, attention cues, importance scores, or layer-wise redundancy to reduce visual tokens~\cite{efficient,efficient2}. While effective for image inputs and generic multimodal settings, these methods primarily focus on spatial redundancy and local token importance, and do not directly address the additional temporal redundancy and cross-frame dependencies introduced by video inputs.

Recent studies have extended token compression to VideoLLMs by incorporating temporal redundancy modeling or adaptive token allocation~\cite{holitom,fastvid,vidcom2,dytok,framefusion,prunevid,adatp,dynamic,multig,prune2drive,keyframe}. For example, FastVID~\cite{fastvid} performs density-based token pruning for video inputs, VidCom2~\cite{vidcom2} ranks tokens according to frame-level and global uniqueness, DyTok~\cite{dytok} dynamically allocates token budgets across frames or segments, PruneVID~\cite{prunevid} distinguishes static and dynamic tokens for compression, and FrameFusion~\cite{framefusion} combines token merging and pruning across model layers during VideoLLM inference. Compared with image-based pruning, video token compression must go beyond redundant token removal to preserve temporal continuity, spatially distinctive content, and global coverage over long video sequences~\cite{keyframe,divprune,prune2drive}. However, most existing methods still emphasize spatial redundancy reduction, segment-level compression, or local temporal heuristics, while the joint modeling of token importance from a global spatio-temporal perspective remains underexplored.

\section{Methodology}
\subsection{Preliminary}
\subsubsection{Video Large Language Models.}
VideoLLMs generally adopt a unified architecture to process multimodal pairs of text and video inputs. 
Given a raw video sequence $X \in \mathbb{R}^{T \times H \times W \times 3}$ and a corresponding textual query, the model processes them through two distinct pathways. The text input is tokenized into a sequence of textual embeddings $H^t$. Whereas the visual input is processed by a pre-trained vision encoder, yielding dense spatial-temporal features $Z \in \mathbb{R}^{T \times M \times D_{vit}}$, where $T$ is the number of sampled frames, $M$ represents the number of spatial patches per frame, and $D_{vit}$ is the feature dimension of the vision encoder. To align the visual representations with the textual semantic space, a projector layer is employed to convert these features into visual tokens. Such operation produces the final visual sequence $H^v \in \mathbb{R}^{T \times N \times D}$, where $N$ denotes the number of retained tokens ($N \le M$), and $D$ matches the hidden dimension of the LLM.
The visual tokens are then flattened into a sequence, concatenated with the text tokens, and fed into the LLM for autoregressive generation of a response sequence $Y$.

\begin{figure*}[t]
    \centering
    \includegraphics[width=0.9\linewidth]{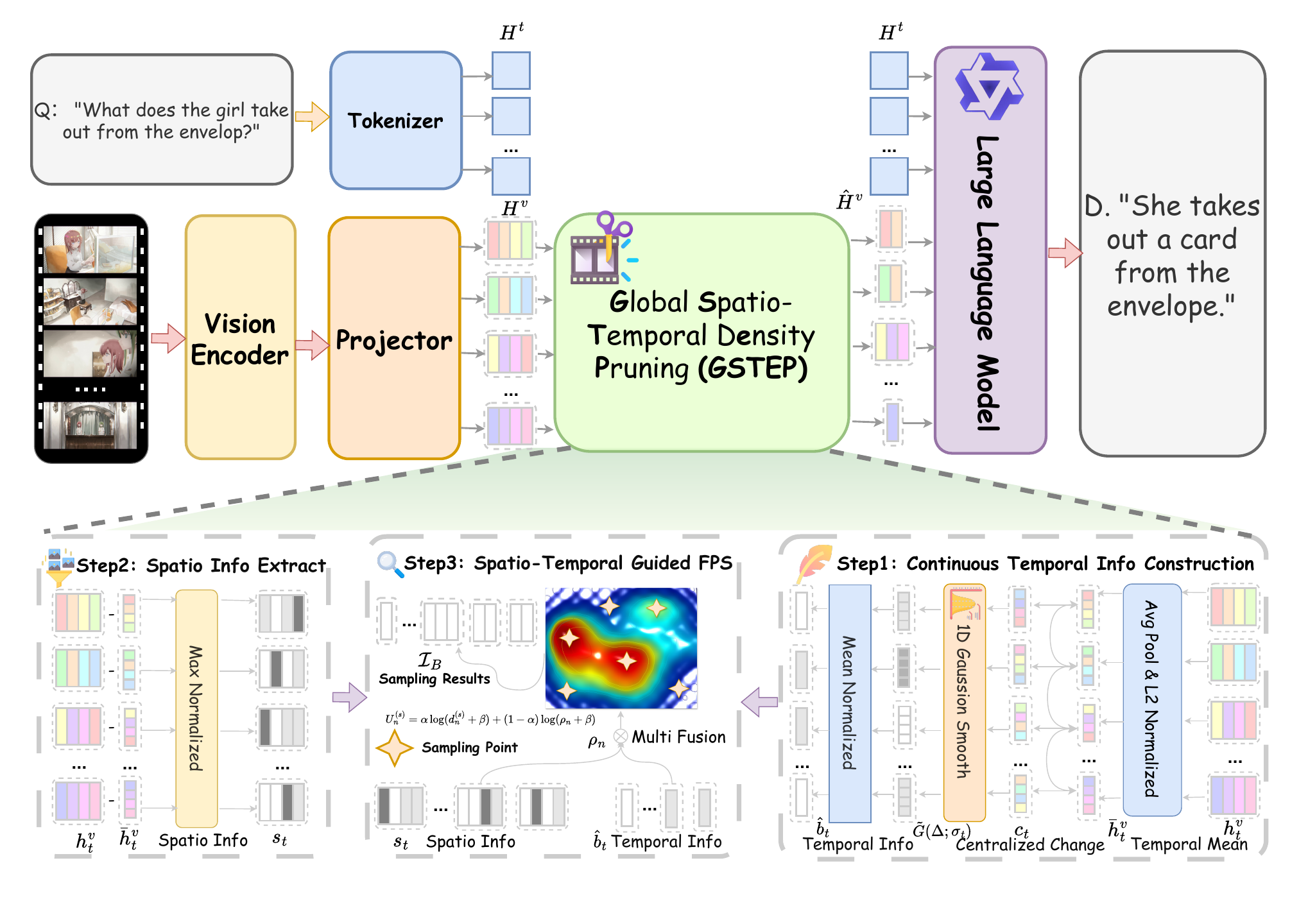}
    \caption{Overview of GSTEP. GSTEP first extracts intra-frame spatial density, then constructs a continuous temporal density field, and finally selects the most representative tokens via Spatio-Temporal guided Farthest Point Sampling (FPS).} 
    \label{fig:pipeline}
\end{figure*}
\subsubsection{Visual Token Pruning.} 
Visual token pruning reduces inference cost by selecting a representative subset $\tilde{H}^v \subset H^v$ under a token budget $B \ll TN$ while preserving the predictive distribution over $Y$:
\begin{equation}
    \tilde{H}^v_{\mathrm{opt}} = \underset{\tilde{H}^v \subset H^v, |\tilde{H}^v| \le B}{\arg\min} \mathcal{L} \Big( P(Y | H^v, H^t) \,\|\, P(Y | \tilde{H}^v, H^t) \Big),
    \label{eq:pruning_objective}
\end{equation}
Here, $\tilde{H}^v_{\mathrm{opt}}$ is the optimal subset of at most $B$ visual tokens. The two distributions correspond to the original and pruned visual sequences, and $\mathcal{L}(\cdot \,\|\, \cdot)$ measures their discrepancy. Eq.~\ref{eq:pruning_objective} states the pruning objective; GSTEP approximates it through training-free density-guided token selection.

\subsection{Global Spatio-Temporal Density Pruning}

\subsubsection{Spatial Density Extraction.}
Tokens corresponding to informative visual content tend to deviate more from the frame-level semantic center than background-like or repetitive tokens. To quantify this spatial distinctiveness, we first compute the semantic center of the $t$-th frame, denoted by $\bar{h}_t^v$, as the average feature of all visual tokens in that frame:
\begin{equation}
    \bar{h}_t^v = \frac{1}{N}\sum_{k=1}^{N} h_{t,k}^v.
\end{equation}
where $h_{t,k}^v$ denotes the feature of the $k$-th visual token in frame $t$. We then measure the deviation of each token from this center and normalize it within the frame. The final spatial density weight $s_{t,i}$ is formulated as:
\begin{equation}
    s_{t,i} = \frac{\left\| h_{t,i}^v - \bar{h}_t^v \right\|_2}{\max\limits_{1 \le j \le N} \left\| h_{t,j}^v - \bar{h}_t^v \right\|_2},
\end{equation}
where a normalized weight $s_{t,i} \in [0, 1]$ closer to $1$ indicates that the token carries highly distinguishable local spatial details relative to its surrounding context.
\subsubsection{Continuous Temporal Density Extraction.}
While spatial density captures the intra-frame token importance, we further evaluate the inter-frame temporal dynamics to extract a continuous temporal density for each frame. 

\textbf{Centered Global Change.} 
We first quantify the temporal transition intensity $c_t$ of the $t$-th frame relative to its local temporal neighborhood. Using the global semantic representation of the frame, the change intensity for intermediate frames $t \in [2, T-1]$ is formulated as the average semantic distance to its adjacent frames:
\begin{equation}
    c_t = \frac{1}{2}d(\bar{h}_{t-1}^v, \bar{h}_t^v) + \frac{1}{2}d(\bar{h}_t^v, \bar{h}_{t+1}^v),
\end{equation}
where $d(\cdot, \cdot)$ is a chosen distance metric (e.g., cosine distance). For the boundary frames, we define $c_1 = d(\bar{h}_1^v, \bar{h}_2^v)$ and $c_T = d(\bar{h}_{T-1}^v, \bar{h}_T^v)$. This yields the raw global change curve $\mathbf{c} = [c_1, \dots, c_T]$.

\textbf{Continuous Smoothing.} 
Directly utilizing the raw change curve $\mathbf{c}$ may introduce abrupt semantic fluctuations. To better reflect the continuous nature of video, we apply a 1D Gaussian smoothing operator over the temporal axis. Given a temporal standard deviation $\sigma_t$ and a truncation radius $r = \lceil 3\sigma_t \rceil$, the normalized Gaussian kernel $\tilde{G}(\Delta; \sigma_t)$ is defined as:
\begin{equation}
    \tilde{G}(\Delta; \sigma_t) = \frac{\exp\left(-\frac{\Delta^2}{2\sigma_t^2}\right)}{\sum_{\delta=-r}^{r} \exp\left(-\frac{\delta^2}{2\sigma_t^2}\right)}.
\end{equation}
where $\Delta \in [-r,r]$ denotes the temporal offset relative to frame $t$. The smoothed frame-level temporal density $\hat{b}_t$ is then obtained by convolving the raw change curve with this kernel:
\begin{equation}
    \hat{b}_t = \sum_{\Delta=-r}^{r} \tilde{G}(\Delta; \sigma_t) \, c_{t+\Delta}.
\end{equation}

\textbf{Mean Normalization.} 
Finally, videos inherently exhibit varying scales of overall action intensities. To align the temporal density scales across different videos and ensure numerical stability, we perform a mean normalization on the smoothed densities:
\begin{equation}
    \hat{b}_t \leftarrow \frac{\hat{b}_t}{\frac{1}{T}\sum_{\tau=1}^{T}\hat{b}_{\tau} + \varepsilon},
\end{equation}
where $\varepsilon$ is a small constant to prevent zero-division. A larger normalized density $\hat{b}_t$ signifies that the $t$-th frame resides in a temporal region with drastic semantic transitions, thus receiving higher priority during token selection.

\subsubsection{Spatio-Temporal Density Pruning.}
Having established the spatial density and continuous temporal density, we now integrate them to guide the final token pruning. To avoid the suboptimal local decisions induced by temporal segmentation, we pool all tokens across the entire video and perform a density-guided global sampling. 

\textbf{Spatio-Temporal Density.} 
We first construct a comprehensive density field by fusing the inter-frame temporal density and intra-frame spatial density. The resulting spatio-temporal density weight for the $i$-th token at the $t$-th frame is defined as:
\begin{equation}
    \rho_{t,i} = \hat{b}_t \cdot s_{t,i}.
\end{equation}
For global selection, we flatten the frame-token index pair $(t,i)$ into a single global index $n \in \{1,\dots,TN\}$, where $n=(t-1)N+i$. Consequently, the density field and the corresponding visual features are denoted as $\rho_n$ and $h_n^v$, respectively.

\textbf{Global Sampling Initialization.} 
Our goal is to select a highly representative subset of $B$ tokens that maximizes both semantic coverage and information density. We achieve this via a density-guided Farthest Point Sampling (FPS) strategy. 
The sampling process is initialized by eagerly selecting the token with the maximum absolute density as the starting point:
\begin{equation}
    j_1 = \arg\max_{n} \rho_n, \qquad \mathcal{I}_1 = \{j_1\},
\end{equation}
where $\mathcal{I}_1$ denotes the index set of selected tokens at the initial step.

\begin{table*}[t]
    \centering
    \renewcommand{\arraystretch}{1.0}
    \setlength{\tabcolsep}{4pt}
    \caption{\textbf{Performance comparison of different pruning methods on LLaVA-OV-7B (LLaVA-OneVision).} \textbf{Rel.} denotes the mean percentage of performance retained relative to the unpruned model, averaged over VideoMME-Overall, LongVideoBench, MLVU, MVBench, and EgoSchema. The best results are \textbf{bolded} and the second-best are \underline{underlined}.}
    \label{tab:main_llava}
    \resizebox{\textwidth}{!}{%
    \begin{tabular}{l cccc ccccc}
        \toprule
        \multirow{2}{*}{\textbf{Method}} & \multicolumn{4}{c}{\textbf{VideoMME}} & \multirow{2}{*}{\textbf{LongVideoBench}} & \multirow{2}{*}{\textbf{MLVU}} & \multirow{2}{*}{\textbf{MVBench}} & \multirow{2}{*}{\textbf{EgoSchema}} & \multirow{2}{*}{\textbf{Rel.}} \\
        \cmidrule(lr){2-5}
        & \textbf{Short} & \textbf{Medium} & \textbf{Long} & \textbf{Overall} & & & & & \\
        \midrule
        \multicolumn{10}{c}{\textit{Original Model (100\%)}} \\
        \rowcolor{oursgray}
        LLaVA-OV-7B & 70.3 & 56.6 & 48.8 & 58.6 & 56.4 & 63.0 & 56.9 & 60.4 & 100.0\% \\
        \midrule
        \multicolumn{10}{c}{\textit{Pruning Ratio ($\downarrow 75\%$)}} \\
        FastV & 65.0 & 53.8 & 47.0 & 55.3 & 53.3 & 59.6 & 55.5 & 57.5 & 95.2\% \\
        FastVID & \underline{69.9} & \underline{56.3} & 47.4 & 57.9 & \underline{56.3} & 61.3 & 56.5 & 59.5 & 98.7\% \\
        CDPruner & 63.2 & 54.8 & 47.1 & 55.0 & 54.3 & 61.2 & 54.7 & 58.3 & 96.0\% \\
        VisionZip & 61.6 & 53.4 & 47.2 & 54.1 & 51.2 & 58.5 & 53.7 & \textbf{60.3} & 94.0\% \\
        VidCom2 & 69.8 & \textbf{56.4} & \underline{49.4} & \underline{58.6} & 54.9 & \underline{62.5} & \textbf{57.2} & \underline{59.7} & \underline{99.2\%} \\
        \rowcolor{oursgreen}
        \textbf{GSTEP (Ours)} & \textbf{70.0} & \textbf{56.4} & \textbf{50.2} & \textbf{58.9} & \textbf{57.2} & \textbf{63.3} & \underline{56.9} & 59.5 & \textbf{100.2\%} \\
        \midrule
        \multicolumn{10}{c}{\textit{Pruning Ratio ($\downarrow 80\%$)}} \\
        FastV & 60.8 & 51.0 & 47.2 & 53.0 & 50.0 & 57.2 & 51.9 & 56.6 & 91.0\% \\
        FastVID & \underline{69.3} & \textbf{56.7} & 47.7 & 57.9 & \underline{56.7} & 61.1 & \underline{56.3} & \underline{59.5} & \underline{98.8\%} \\
        CDPruner & 62.2 & 52.7 & 46.9 & 53.9 & 53.3 & 59.7 & 53.9 & 57.2 & 94.1\% \\
        VisionZip & 60.8 & 51.0 & 47.1 & 53.0 & 50.0 & 57.1 & 53.0 & \textbf{59.8} & 92.4\% \\
        VidCom2 & \textbf{69.4} & \underline{56.4} & \underline{48.0} & \underline{58.0} & 53.7 & \underline{61.9} & \textbf{56.6} & \textbf{59.8} & 98.2\% \\
        \rowcolor{oursgreen}
        \textbf{GSTEP (Ours)} & 68.9 & \underline{56.4} & \textbf{49.8} & \textbf{58.4} & \textbf{56.9} & \textbf{63.4} & \textbf{56.6} & 59.1 & \textbf{99.7\%} \\
        \midrule
        \multicolumn{10}{c}{\textit{Pruning Ratio ($\downarrow 85\%$)}} \\
        FastV & 58.6 & 50.8 & 47.1 & 52.1 & 47.7 & 56.6 & 50.5 & 55.6 & 88.8\% \\
        FastVID & \textbf{69.7} & \textbf{55.8} & 47.7 & \textbf{57.7} & \underline{56.5} & \underline{60.9} & \underline{56.0} & 58.8 & \underline{98.2\%} \\
        CDPruner & 61.9 & 52.2 & 45.8 & 53.3 & 52.4 & 58.8 & 52.9 & 57.0 & 92.9\% \\
        VisionZip & 50.4 & 45.8 & 41.8 & 46.0 & 46.9 & 54.4 & 50.3 & \textbf{59.8} & 87.1\% \\
        VidCom2 & 65.8 & \underline{54.8} & \underline{48.1} & 56.2 & 52.0 & 58.9 & 54.3 & \underline{59.2} & 95.0\% \\
        \rowcolor{oursgreen}
        \textbf{GSTEP (Ours)} & \underline{69.0} & \underline{54.8} & \textbf{49.0} & \underline{57.6} & \textbf{57.0} & \textbf{63.5} & \textbf{57.0} & 58.9 & \textbf{99.6\%} \\
        \midrule
        \multicolumn{10}{c}{\textit{Pruning Ratio ($\downarrow 90\%$)}} \\
        FastV & 54.1 & 49.8 & 44.6 & 49.4 & 46.8 & 55.6 & 49.8 & 54.6 & 86.7\% \\
        FastVID & \underline{67.4} & \textbf{56.0} & \textbf{48.6} & \textbf{57.2} & \textbf{55.6} & \underline{60.5} & \underline{55.9} & \textbf{58.6} & \underline{97.5\%} \\
        CDPruner & 59.4 & 51.7 & 45.4 & 52.2 & 49.7 & 57.7 & 51.1 & 55.8 & 90.2\% \\
        VisionZip & 50.4 & 45.8 & 41.8 & 46.0 & 43.5 & 51.5 & 44.4 & \underline{58.0} & 82.3\% \\
        VidCom2 & 61.6 & 52.0 & \underline{47.7} & 53.7 & 50.3 & 58.1 & 52.6 & 57.5 & 92.1\% \\
        \rowcolor{oursgreen}
        \textbf{GSTEP (Ours)} & \textbf{67.8} & \underline{55.3} & 47.1 & \underline{56.7} & \underline{55.2} & \textbf{62.1} & \textbf{56.1} & \textbf{58.6} & \textbf{97.8\%} \\
        \bottomrule
    \end{tabular}%
    }
\end{table*}

\begin{table*}[t]
    \centering
    \renewcommand{\arraystretch}{1.0}
    \setlength{\tabcolsep}{4pt}
    \caption{\textbf{Performance comparison of different pruning methods on Qwen2.5-VL-7B.} \textbf{Rel.} denotes the mean percentage of performance retained. The best results are \textbf{bolded} and the second-best are \underline{underlined}.}
    \label{tab:main_qwen}
    \resizebox{\textwidth}{!}{%
    \begin{tabular}{l cccc ccccc}
        \toprule
        \multirow{2}{*}{\textbf{Method}} & \multicolumn{4}{c}{\textbf{VideoMME}} & \multirow{2}{*}{\textbf{LongVideoBench}} & \multirow{2}{*}{\textbf{MLVU}} & \multirow{2}{*}{\textbf{MVBench}} & \multirow{2}{*}{\textbf{EgoSchema}} & \multirow{2}{*}{\textbf{Rel.}} \\
        \cmidrule(lr){2-5}
        & \textbf{Short} & \textbf{Medium} & \textbf{Long} & \textbf{Overall} & & & & & \\
        \midrule
        \multicolumn{10}{c}{\textit{Original Model (100\%)}} \\
        \rowcolor{oursgray}
        Qwen2.5-VL-7B & 72.8 & 61.1 & 50.2 & 61.4 & 58.5 & 59.3 & 68.5 & 58.2 & 100.0\% \\
        \midrule
        \multicolumn{10}{c}{\textit{Pruning Ratio ($\downarrow 75\%$)}} \\
        FastV & 68.7 & 56.4 & 51.0 & 58.7 & 54.4 & 56.5 & 65.6 & 56.0 & 95.2\% \\
        FastVID & \underline{71.2} & \textbf{57.8} & 49.9 & 59.6 & \underline{58.0} & 58.1 & 65.5 & \underline{56.7} & 97.4\% \\
        CDPruner & 70.2 & 56.7 & 49.8 & 58.9 & 57.4 & \underline{58.6} & 66.3 & 55.8 & 97.1\% \\
        VisionZip & 71.0 & \underline{56.9} & \textbf{51.3} & \underline{59.7} & 57.2 & 58.0 & \textbf{67.1} & \underline{56.7} & \underline{97.6\%} \\
        VidCom2 & 69.3 & 55.6 & \underline{51.2} & 58.7 & 55.2 & 58.0 & \underline{66.5} & \textbf{56.8} & 96.5\% \\
        \rowcolor{oursgreen}
        \textbf{GSTEP (Ours)} & \textbf{71.7} & 56.7 & 51.1 & \textbf{59.8} & \textbf{58.8} & \textbf{58.9} & 65.2 & 55.8 & \textbf{97.7\%} \\
        \midrule
        \multicolumn{10}{c}{\textit{Pruning Ratio ($\downarrow 80\%$)}} \\
        FastV & 68.1 & 55.4 & \textbf{51.7} & 58.4 & 53.5 & 55.3 & 64.5 & 55.9 & 94.0\% \\
        FastVID & \underline{69.9} & 56.3 & 49.7 & 58.6 & \underline{57.8} & 57.9 & 64.7 & \textbf{56.4} & 96.6\% \\
        CDPruner & 68.3 & \underline{57.2} & 50.3 & 58.6 & 57.1 & \underline{58.3} & 65.4 & 55.3 & 96.4\% \\
        VisionZip & \underline{69.9} & \textbf{57.6} & 49.1 & \underline{58.9} & 56.7 & 58.1 & \textbf{66.3} & \underline{56.2} & \underline{96.8\%} \\
        VidCom2 & 68.1 & 54.9 & 49.4 & 57.5 & 55.2 & 57.7 & \underline{65.8} & \underline{56.2} & 95.6\% \\
        \rowcolor{oursgreen}
        \textbf{GSTEP (Ours)} & \textbf{70.2} & 56.8 & \underline{51.1} & \textbf{59.4} & \textbf{59.1} & \textbf{58.8} & 64.8 & 55.2 & \textbf{97.3\%} \\
        \midrule
        \multicolumn{10}{c}{\textit{Pruning Ratio ($\downarrow 85\%$)}} \\
        FastV & 66.4 & 54.6 & 50.2 & 57.1 & 52.5 & 53.4 & 62.8 & 55.3 & 91.9\% \\
        FastVID & 68.2 & \textbf{56.9} & 49.4 & \underline{58.2} & \underline{56.8} & 56.9 & 63.6 & \textbf{56.0} & 95.4\% \\
        CDPruner & 68.1 & \underline{55.9} & \underline{50.3} & 58.1 & 55.7 & \textbf{58.1} & \underline{65.2} & 54.7 & 95.4\% \\
        VisionZip & \underline{68.3} & \underline{55.9} & 48.6 & 57.6 & 56.0 & 57.1 & \textbf{65.6} & \underline{55.9} & \underline{95.5\%} \\
        VidCom2 & 67.4 & 54.9 & 48.6 & 57.0 & 54.5 & 55.8 & 64.6 & 55.6 & 94.0\% \\
        \rowcolor{oursgreen}
        \textbf{GSTEP (Ours)} & \textbf{69.1} & 55.4 & \textbf{50.4} & \textbf{58.3} & \textbf{58.0} & \underline{57.8} & 64.0 & 54.5 & \textbf{95.7\%} \\
        \midrule
        \multicolumn{10}{c}{\textit{Pruning Ratio ($\downarrow 90\%$)}} \\
        FastV & 63.8 & 53.3 & \underline{49.2} & 55.4 & 50.1 & 50.8 & 61.1 & 54.1 & 88.7\% \\
        FastVID & \underline{66.3} & 53.6 & 49.0 & \underline{56.3} & \underline{55.4} & 55.8 & 62.3 & \textbf{55.6} & \underline{93.4\%} \\
        CDPruner & 65.4 & 53.4 & \textbf{49.6} & 56.1 & 55.1 & \underline{56.5} & \textbf{63.9} & 53.5 & 93.2\% \\
        VisionZip & 65.8 & \textbf{54.0} & 47.9 & 55.9 & 53.6 & 56.1 & \underline{63.7} & \underline{54.7} & 92.8\% \\
        VidCom2 & 65.7 & 53.6 & 48.6 & 55.9 & 51.8 & 54.1 & 61.5 & 53.4 & 90.5\% \\
        \rowcolor{oursgreen}
        \textbf{GSTEP (Ours)} & \textbf{66.8} & \underline{53.7} & 48.9 & \textbf{56.4} & \textbf{58.1} & \textbf{57.2} & 62.9 & 53.7 & \textbf{94.3\%} \\
        \bottomrule
    \end{tabular}%
    }
\end{table*}

\textbf{Log-Decoupled Iterative Sampling.} 
Standard FPS~\cite{divprune,prune2drive} purely relies on feature distances, which risks sampling isolated but uninformative outlier tokens. To strictly prioritize high-density regions while maintaining global semantic diversity, we introduce a log-decoupled scoring mechanism. At the $s$-th iterative step, let $\mathcal I_s$ denote the selected token set, where $|\mathcal I_s|=s<B$. For each unselected token $n \notin \mathcal{I}_s$, its shortest cosine distance to the already selected set is formally defined as:
\begin{equation}
    l_n^{(s)} = \min_{j \in \mathcal{I}_s} \left( 1 - (\tilde{h}_n^v)^\top \tilde{h}_j^v \right).
\end{equation}
We then formulate a composite selection score that balances the representation distance and information density using a logarithmic decoupling function:
\begin{equation}
    U_n^{(s)} = \alpha \log(l_n^{(s)} + \beta) + (1-\alpha)\log(\rho_n + \beta),
\label{eq:utility_score}
\end{equation}
where $\alpha \in [0, 1]$ is a trade-off hyperparameter, and $\beta$ is a small constant to ensure numerical stability. The logarithmic projection helps align the scales of the distance term and the density term.
The token exhibiting the highest composite score is subsequently added to the selected set:
\begin{equation}
    j_{s+1} = \arg\max_{n \notin \mathcal{I}_s} U_n^{(s)}, \qquad \mathcal{I}_{s+1} = \mathcal{I}_s \cup \{j_{s+1}\}, \quad s=1,\dots,B-1.
\end{equation}
\section{Experiments}
\subsection{Experimental Setup}

\textbf{Benchmarks.}
We evaluate GSTEP on five widely used video understanding benchmarks: VideoMME~\cite{videomme}, LongVideoBench~\cite{longvideobench}, MLVU~\cite{mlvu}, MVBench~\cite{mvbench}, and EgoSchema~\cite{egoschema}. These benchmarks cover diverse video durations and evaluation scenarios, providing a comprehensive testbed for evaluating both the effectiveness and the generalization capability of GSTEP. For all benchmarks, we follow their default evaluation protocols and metrics.

\textbf{Backbone models and baselines.}
We conduct the main experiments on two representative open-source VideoLLMs, LLaVA-OV-7B~\cite{llava_ov} and Qwen2.5-VL-7B~\cite{qwen2_5_vl}. We further include supplementary evaluation on Qwen3-VL-8B~\cite{qwen3_vl} at a 75\% pruning ratio to examine the transferability of GSTEP to newer backbones. For comparison, we consider five recent visual token pruning baselines in the main experiments: FastV~\cite{fastv}, FastVID~\cite{fastvid}, CDPruner~\cite{cdpruner}, VisionZip~\cite{visionzip}, and VidCom2~\cite{vidcom2}.

\textbf{Implementation details.}
We report results under four pruning ratios: 75\%, 80\%, 85\%, and 90\%, corresponding to retaining 25\%, 20\%, 15\%, and 10\% of the visual tokens, respectively. Unless otherwise specified, GSTEP uses $\sigma_t=1.0$ and $\alpha=0.5$, and all main experiments share the same hyperparameter setting across datasets and pruning ratios. All methods are evaluated under the same video preprocessing, prompt format, decoding configuration, and token budget settings. Evaluations are conducted using the LMMs-Eval framework~\cite{lmms} on 6 $\times$ A100 GPUs with 40 GB memory each, with FlashAttention2~\cite{flashattention2} enabled whenever applicable. Since the original FastV implementation is incompatible with FlashAttention2, we adapt it by decoupling token-importance estimation into a lightweight auxiliary step while preserving its original pruning logic. Supplementary evaluation on Qwen3-VL-8B is conducted at a 75\% pruning ratio.
\subsection{Main Results}
\textbf{Overall comparison.}
Tables~\ref{tab:main_llava} and~\ref{tab:main_qwen} compare GSTEP with recent token pruning baselines on LLaVA-OV-7B and Qwen2.5-VL-7B across four pruning ratios. Overall, GSTEP achieves strong and stable performance across diverse benchmarks, with increasingly pronounced advantages under higher compression, especially on long-video benchmarks such as LongVideoBench and MLVU. This trend supports modeling videos as a continuous spatio-temporal information flow and performing global rather than discrete local token selection.

\textbf{Results on LLaVA-OV-7B.}
As shown in Table~\ref{tab:main_llava}, GSTEP achieves the best average performance retention on LLaVA-OV-7B under all four pruning ratios, reaching 100.2\%, 99.7\%, 99.6\%, and 97.8\% from 75\% to 90\% pruning, respectively. At 85\% pruning, it attains the best results on LongVideoBench, MLVU, and MVBench, while remaining among the top methods on VideoMME. At 90\% pruning, GSTEP retains 97.8\% of the original average performance and leads all baselines in relative retention, demonstrating more graceful degradation as the token budget becomes constrained.

A closer look shows that the gains of GSTEP are especially evident on temporally demanding benchmarks. Compared with the strongest competing baselines, GSTEP consistently performs better on MLVU across all pruning ratios and achieves stronger or comparable performance on LongVideoBench in most settings. On VideoMME, GSTEP is particularly advantageous on the long subset, suggesting that global spatio-temporal density modeling preserves key information over longer temporal contexts. Although some baselines remain competitive on shorter or less temporally intensive tasks, GSTEP is the most stable top-performing method overall, indicating better preservation of temporally critical evidence under aggressive pruning.

\textbf{Results on Qwen2.5-VL-7B.}
Table~\ref{tab:main_qwen} shows a similar trend on Qwen2.5-VL-7B: GSTEP achieves the best average retention under all four pruning ratios, preserving 97.7\%, 97.3\%, 95.7\%, and 94.3\% of the original performance from 75\% to 90\% pruning, respectively. It also delivers the best overall VideoMME and LongVideoBench scores across all four settings, supporting its generalization across VideoLLM backbones.
The gains remain clear on benchmarks requiring temporal reasoning and long-range video understanding. At 90\% pruning, GSTEP leads on MLVU and LongVideoBench. At 85\% pruning, it ranks first on LongVideoBench and remains competitive on MLVU. Although it sometimes trails the strongest baseline on MVBench and EgoSchema, GSTEP maintains the best average retention at every compression ratio, offering a robust trade-off between pruning aggressiveness and performance preservation.

\begin{figure}[t]
    \centering
    \includegraphics[width=\columnwidth]{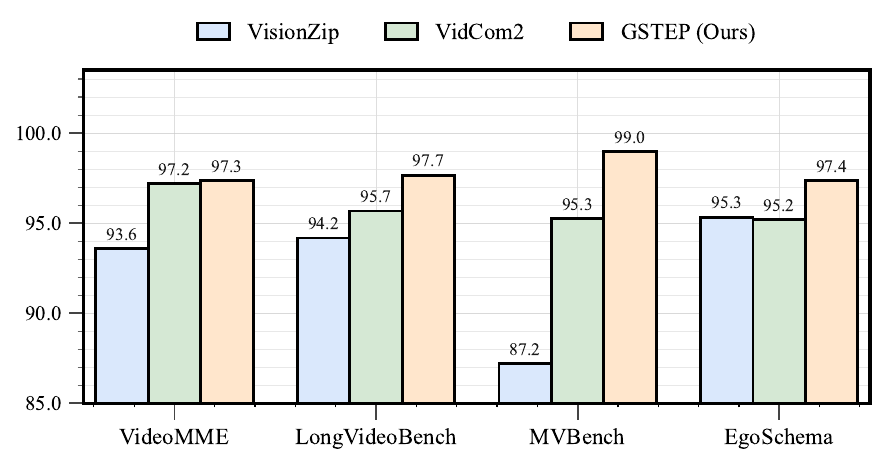}
    \caption{Relative performance on Qwen3-VL-8B at 75\% pruning ratio. All values are normalized to the unpruned baseline, where 100 denotes the original performance.}
    \label{fig:qwen3_relperf}
\end{figure}

\textbf{Supplementary results on Qwen3-VL-8B.}
Figure~\ref{fig:qwen3_relperf} reports relative performance at 75\% pruning on Qwen3-VL-8B. GSTEP achieves the best retention on all four benchmarks, reaching 97.3\% on VideoMME, 97.7\% on LongVideoBench, 99.0\% on MVBench, and 97.4\% on EgoSchema. Compared with VisionZip and VidCom2, it shows the clearest advantage on MVBench while remaining stronger on the other benchmarks, supporting transfer to newer VideoLLM backbones. Taken together, the results across three backbones show strong average retention and particularly clear gains on temporally demanding benchmarks, supporting global spatio-temporal token selection under aggressive compression.

\begin{table}[t]
    \centering
    \scriptsize
    \setlength{\tabcolsep}{2.5pt}
    \renewcommand{\arraystretch}{0.92}
    \caption{Efficiency comparison on LLaVA-OV-7B over LongVideoBench at 75\% and 85\% pruning ratios. Parentheses show relative change with respect to the original model.}
    \label{tab:efficiency}
    \resizebox{\columnwidth}{!}{
    \begin{tabular}{lcccc}
        \toprule
        Method & E2E (s)$\downarrow$ & LLM Time (s)$\downarrow$ & GPU (GB)$\downarrow$ & KV (MB)$\downarrow$ \\
        \midrule
        \rowcolor{oursgray}
        Original (100\%) & 907.5 & 280.3 & 17.57 & 340 \\
        \midrule
        \multicolumn{5}{c}{\textit{Pruning Ratio ($\downarrow 75\%$)}} \\
        \midrule
        FastV & 805.5 \dec{11.2} & 184.1 \dec{34.3} & 16.50 \dec{6.1} & 110 \dec{67.6} \\
        FastVID & 790.6 \dec{12.9} & 173.0 \dec{38.3} & 16.00 \dec{8.9} & 90 \dec{73.5} \\
        CDPruner & 961.8 \inc{6.0} & 338.9 \inc{20.9} & 16.10 \dec{8.4} & 90 \dec{73.5} \\
        VidCom2 & 767.1 \dec{15.5} & 167.4 \dec{40.3} & 16.00 \dec{8.9} & 90 \dec{73.5} \\
        \rowcolor{oursgreen}
        \textbf{GSTEP (Ours)} & 777.5 \dec{14.3} & 173.6 \dec{38.1} & 16.00 \dec{8.9} & 90 \dec{73.5} \\
        \midrule
        \multicolumn{5}{c}{\textit{Pruning Ratio ($\downarrow 85\%$)}} \\
        \midrule
        FastV & 774.3 \dec{14.7} & 169.5 \dec{39.5} & 16.48 \dec{6.2} & 77 \dec{77.4} \\
        FastVID & 765.6 \dec{15.6} & 156.2 \dec{44.3} & 15.80 \dec{10.1} & 55 \dec{83.8} \\
        CDPruner & 874.3 \dec{3.7} & 258.0 \dec{7.9} & 15.80 \dec{10.1} & 55 \dec{83.8} \\
        VidCom2 & 762.0 \dec{16.0} & 152.4 \dec{45.6} & 15.80 \dec{10.1} & 55 \dec{83.8} \\
        \rowcolor{oursgreen}
        \textbf{GSTEP (Ours)} & 754.8 \dec{16.8} & 157.7 \dec{43.7} & 15.80 \dec{10.1} & 55 \dec{83.8} \\
        \bottomrule
    \end{tabular}
    }
\end{table}

\begin{table}[ht]
    \centering
    \small
    \setlength{\tabcolsep}{3pt}
    \renewcommand{\arraystretch}{1.05}
    \caption{Ablation study of GSTEP on LLaVA-OV-7B at an 85\% pruning ratio. VideoMME reports the overall score.}
    \label{tab:ablation}
    \resizebox{\columnwidth}{!}{
    \begin{tabular}{lcccc}
        \toprule
        \abhead{1.8cm}{\textbf{Method}} & \abhead{1.5cm}{\textbf{VideoMME}} & \abhead{1.8cm}{\textbf{LongVideo\\Bench}} & \abhead{1.0cm}{\textbf{MLVU}} & \abhead{0.8cm}{\textbf{Avg.}} \\
        \midrule
        \multicolumn{5}{l}{\textbf{Temporal construction}} \\
        \midrule
        Raw change & 57.5 & 55.5 & 63.3 & 58.8 \\
        Fixed smoothing & \textbf{57.7} & 56.4 & 63.2 & 59.1 \\
        \rowcolor{oursgreen}
        Gaussian smoothing & 57.6 & \textbf{57.0} & \textbf{63.5} & \textbf{59.4} \\
        \midrule
        \multicolumn{5}{l}{\textbf{Density fusion}} \\
        \midrule
        Temporal only & 55.8 & 54.1 & 61.9 & 57.3 \\
        Spatial only & 56.6 & 56.7 & 63.3 & 58.9 \\
        \rowcolor{oursgreen}
        Full GSTEP & \textbf{57.6} & \textbf{57.0} & \textbf{63.5} & \textbf{59.4} \\
        \midrule
        \multicolumn{5}{l}{\textbf{Sampling strategy}} \\
        \midrule
        Random & 56.4 & 55.9 & 61.6 & 58.0 \\
        Uniform & 56.3 & 55.8 & 61.7 & 57.9 \\
        Top-$k$ only & 55.0 & 55.4 & 61.3 & 57.2 \\
        FPS only & 55.3 & 55.3 & 62.8 & 57.8 \\
        \rowcolor{oursgreen}
        Density-guided FPS & \textbf{57.6} & \textbf{57.0} & \textbf{63.5} & \textbf{59.4} \\
        \bottomrule
    \end{tabular}
    }
\end{table}

\begin{figure*}[t]
    \centering
    \includegraphics[width=\textwidth]{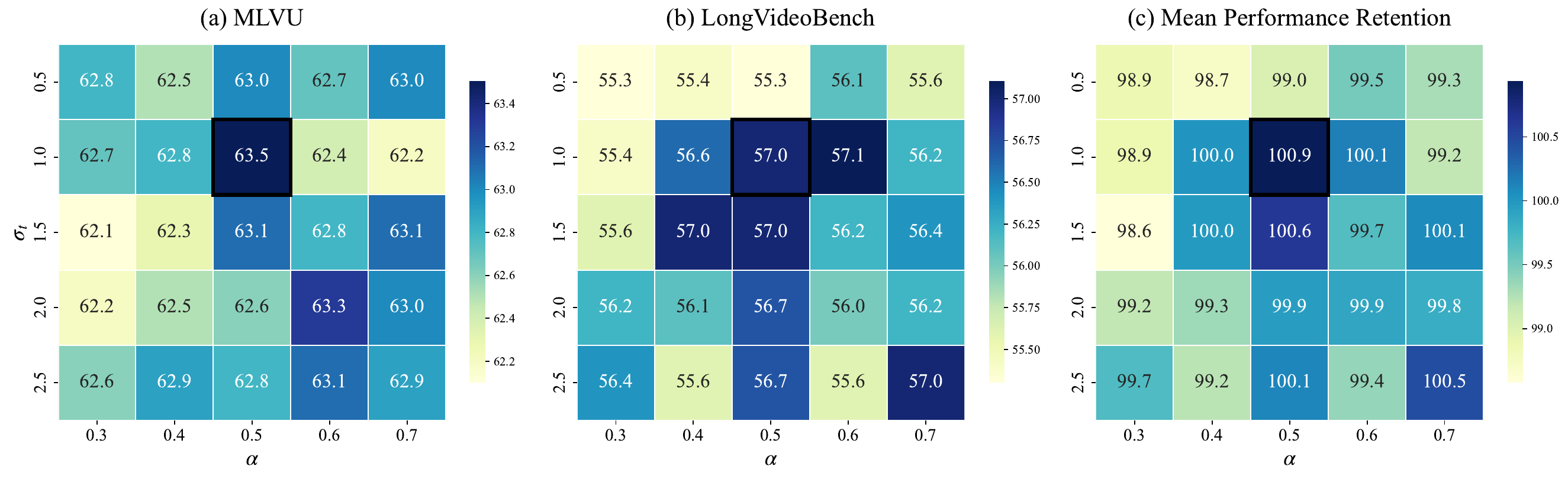}
    \caption{Parameter sensitivity of GSTEP on LLaVA-OV-7B under an 85\% pruning ratio. Heatmaps show the benchmark scores on (a) MLVU and (b) LongVideoBench, as well as (c) the mean performance retention averaged over the two benchmarks. The default setting $(\sigma_t,\alpha)=(1.0,0.5)$ is highlighted by a black box.}
    \label{fig:param_sensitivity}
\end{figure*}
\subsection{Efficiency Analysis}
Table~\ref{tab:efficiency} compares GSTEP with representative pruning baselines on LongVideoBench at 75\% and 85\% pruning. At 75\% pruning, GSTEP reduces E2E latency from 907.5\,s to 777.5\,s (\textbf{1.17$\times$} speedup) and LLM inference time to 173.6\,s, while matching the lowest GPU memory and KV cache usage. Although VidCom2 has slightly lower E2E latency at this ratio, GSTEP remains competitive while maintaining strong accuracy retention. At 85\% pruning, GSTEP achieves the best E2E latency of 754.8\,s (\textbf{1.20$\times$} speedup), reduces LLM time to 157.7\,s, and matches the lowest memory footprint of 15.80\,GB GPU memory and 55\,MB KV cache. Together, these results demonstrate a practical efficiency--performance trade-off for VideoLLMs.

\subsection{Ablation Study}
We examine three key GSTEP components: temporal density construction, spatio-temporal density fusion, and density-guided global sampling. All ablations use LLaVA-OV-7B with the default hyperparameters ($\sigma_t=1.0$, $\alpha=0.5$) at an 85\% pruning ratio. We report VideoMME (overall), LongVideoBench, and MLVU scores.

Table~\ref{tab:ablation} reports the results. Temporal density construction has a noticeable impact: both smoothing variants improve over the raw change curve overall, especially on LongVideoBench. Fixed smoothing achieves the best VideoMME score, whereas Gaussian smoothing performs best on LongVideoBench and MLVU and attains the strongest average. Thus, smoothing provides a more stable temporal importance signal, although the two variants differ only slightly.

Second, both temporal density and spatial density are important for effective token pruning. Using only the temporal branch or only the spatial branch leads to weaker results than the full model on all three benchmarks. This indicates that effective video token pruning requires not only identifying informative temporal regions, but also distinguishing spatially salient visual tokens across the video. Combining the two signals yields the strongest overall performance.

Third, the final sampling strategy also plays an important role. Random and uniform sampling provide weaker baselines, indicating that neither unguided selection nor simple uniform coverage is sufficient. Likewise, selecting tokens solely by density (\textit{Top-$k$ only}) or solely by feature coverage (\textit{FPS only}) further degrades performance. In contrast, density-guided FPS achieves the best results on all three benchmarks, confirming the importance of jointly considering token density and global coverage during sampling.

\subsection{Parameter Sensitivity Analysis}
We analyze the sensitivity of GSTEP to the temporal Gaussian smoothing scale $\sigma_t$ and the density--coverage trade-off coefficient $\alpha$ in Eq.~\ref{eq:utility_score}. Experiments use LLaVA-OV-7B at an 85\% pruning ratio and report results on MLVU and LongVideoBench.
As shown in Figure~\ref{fig:param_sensitivity}, GSTEP remains stable across the grid: score variation is within 1.4 points on MLVU and 1.8 points on LongVideoBench, while mean performance retention remains high over a broad region. Thus, GSTEP is robust to $\sigma_t$ and $\alpha$ rather than relying on a narrow single-point optimum.
In particular, the strongest region lies around moderate temporal smoothing and a balanced density--coverage trade-off, especially near $\sigma_t=1.0$--$1.5$ and $\alpha=0.5$--$0.6$. Although the exact optimum varies slightly across benchmarks, the default setting $(\sigma_t,\alpha)=(1.0,0.5)$ lies within this stable high-performance region and achieves the best mean performance retention. We therefore use it as the default configuration in all main experiments.

\section{Conclusion}
In this paper, we proposed GSTEP, a global spatio-temporal density pruning framework for VideoLLMs.
By modeling video as a continuous spatio-temporal information flow and performing global token sampling based on unified temporal density and spatial density, GSTEP effectively preserves critical semantic information under limited token budgets.
Extensive experiments show that GSTEP consistently achieves strong accuracy-efficiency trade-offs across multiple VideoLLMs and benchmarks, demonstrating its effectiveness and generalization ability for efficient video token compression.

\begin{acks}
This work was supported by the National Key R\&D Program of China
(No.~2024YDLN0004) and the Fundamental Research Funds for the Central
Universities (No.~WK2102026004).
\end{acks}

\bibliographystyle{ACM-Reference-Format}
\bibliography{ref}

\end{document}